\documentclass[letterpaper]{article} % DO NOT CHANGE THIS
\usepackage{aaai2027}  % DO NOT CHANGE THIS
\usepackage[hyphens]{url}  % DO NOT CHANGE THIS
\usepackage{booktabs, multirow, xcolor, graphicx, amssymb} % DO NOT CHANGE THIS
\usepackage{natbib}  % DO NOT CHANGE THIS AND DO NOT ADD ANY OPTIONS TO IT
\usepackage{caption, subcaption} % DO NOT CHANGE THIS AND DO NOT ADD ANY OPTIONS TO IT
\usepackage{algorithm}
\usepackage{algorithmic}

\usepackage{newfloat}
\usepackage{listings}
\DeclareCaptionStyle{ruled}{labelfont=normalfont,labelsep=colon,strut=off} % DO NOT CHANGE THIS
\floatstyle{ruled}
\newfloat{listing}{tb}{lst}{}
\floatname{listing}{Listing}

\usepackage{amsmath}
\title{Rethinking Text-Based Image Retrieval in Specific Domain}
\author{
    Jingyang Tan\textsuperscript{1}\thanks{Email: tanjingyang.wjf@gmail.com}\equalcontrib,
    Sheng Yang\textsuperscript{2}\equalcontrib,
    Yuanpeng Chen\textsuperscript{4},
    Jian Wang\textsuperscript{2}\corresponding,
    Nianjin Ye\textsuperscript{3},\\
    Chen Xing\textsuperscript{3},
    Lanpeng Jia\textsuperscript{3}\corresponding
}
\affiliations{
\textsuperscript{1}Department of Electronic and Communication Engineering, Harbin Institute of Technology\\
\textsuperscript{2}School of Data Science, Fudan University,\quad
\textsuperscript{3}Changhong Intelligent Robot,\quad
\textsuperscript{4}Independent Researcher\\
}

\makeatletter
\gdef\copyright@on{}
\makeatother

\begin{document}

\maketitle

\begin{abstract}
Driven by the rapid advancement of vision-language representation learning, Text-based Image Retrieval (TBIR) has made notable progress. However, existing benchmarks are predominantly constructed on an exclusive single-match assumption between query and images. While effective in general scenarios, this assumption fails to reflect practical system performance in specific domains (e.g., surveillance), where a single query often corresponds to multiple relevant candidate images. To address this limitation, we design a Domain-Specific Multi-Match Text-based Image Retrieval (DSMM-TBIR) data engine. Leveraging this engine, we construct Security Multi-Match TBIR (SecMM-TBIR), a benchmark comprising 50k surveillance images with 200 comprehensive queries. Furthermore, we observe that vanilla contrastive learning in specific domains suffers from severe false negatives, forcing the model to push apart semantically similar pairs and thus degrading retrieval performance. We propose the Semantic-Aware Fine-Tuning (SAFT) framework to address semantic compression in specific domains, which incorporates Semantic-Aware Soft-Label Supervision (SASS) and Intra-modal Structural Distillation (ISD) to establish a promising paradigm for domain-specific TBIR tasks. Experiments across diverse CLIP-like models demonstrate that SAFT yields an average mAP@20 gain of 7.8 points on SecMM-TBIR over standard image-text contrastive (ITC) fine-tuning, while also improving general-domain performance. The entire benchmark will be released to facilitate further research.
\end{abstract}

\section{Introduction}

Text-based Image Retrieval (TBIR), which retrieves relevant images from large-scale databases given a textual query, has emerged as a fundamental vision-language task. This multi-modal synergy has powered wide spectrum of real-world applications, yielding substantial benefits for scenarios where the ability to rapidly and accurately retrieve targeted visual instances from massive galleries is the core requirement.

\begin{figure}[!t]
    \centering
    \includegraphics[width=1.0\linewidth]{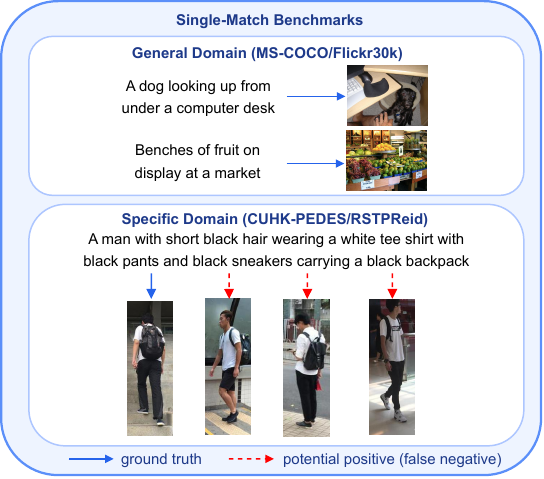}
    \caption{\textbf{Single-Match Benchmarks.} 
    The single-match paradigm ignores multiple potential matches, leading to biased training and evaluation, particularly in specific domain.}
    \label{fig:benchmark_compare}
\end{figure}

The development of robust benchmarks is fundamental to the advancement of TBIR. In the general domain, Flickr30K~\citep{flickrentitiesijcv} and MS-COCO~\citep{lin2014microsoft} serve as the predominant testbeds, characterized by their broad semantic coverage, providing an excellent proxy for evaluating generalized cross-modal alignment capabilities. However,  these benchmarks are predominantly based on the single-match query-image mapping paradigm. While highly effective in general domains, this assumption fundamentally contradicts the practical requirements of specific scenarios. As illustrated in Fig~\ref{fig:benchmark_compare}, in such compressed semantic domain, textual descriptions tend to be simple and often naturally correspond to multiple relevant visual instances. Evaluating models with single-match labels in specific domains inherently biases standard metrics by treating potential positives as negatives, preventing them from reflecting actual retrieval performance. To facilitate the rapid construction of benchmarks across diverse vertical scenarios for evaluation, we design the \textbf{Domain-Specific Multi-Match Text-based Image Retrieval} (DSMM-TBIR) data engine. It leverages the generative and comprehension capabilities of Large Language Models (LLMs) and Vision-Language Models (VLMs), complemented by verification from multiple expert universal multi-modal embedding models. Using this engine, we construct \textbf{Security Multi-Match TBIR} (SecMM-TBIR), a multi-match benchmark comprising 50k surveillance images and 200 comprehensive queries.

\begin{figure}[!t]
    \centering
    \includegraphics[width=1.0\linewidth]{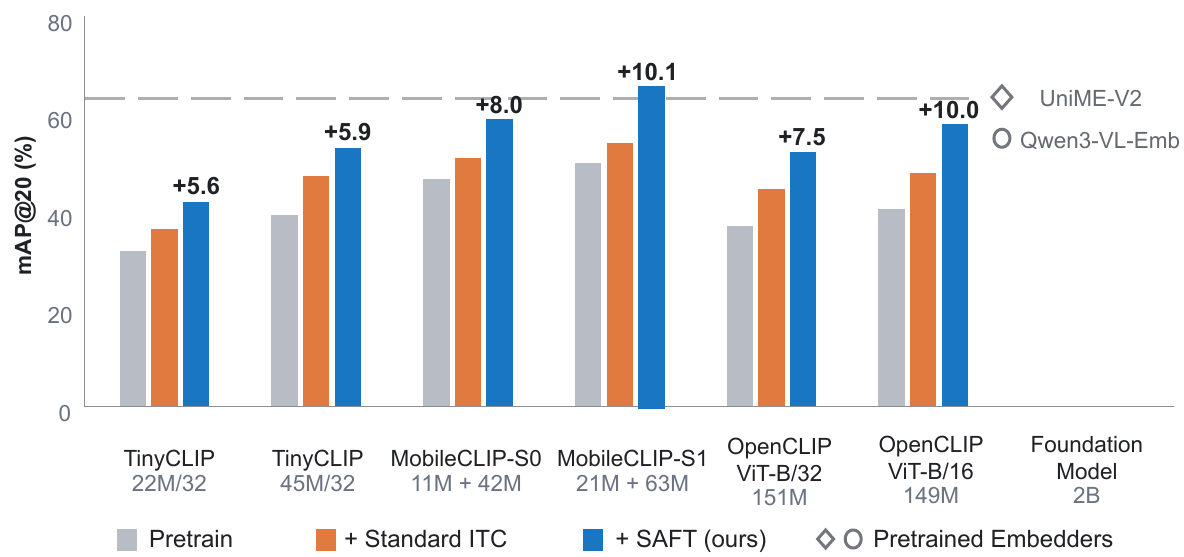}
    \caption{\textbf{Consistent Improvement on SecMM-TBIR Benchmark Driven by SAFT.} It yields an average gain of $+7.8$ mAP@20 over standard ITC fine-tuning, allowing MobileCLIP-S1 to outperform 2B pretrained embedders.}
    \label{fig:first}
\end{figure}

In the context of multi-modal contrastive learning, the prevalence of false negatives (FNs) poses a critical bottleneck. Existing research has primarily diverged into two distinct paradigms: hard negative mining (HNM) and soft-label supervision. HNM is widely used to identify informative hard negatives (HNs) for enhanced contrastive learning~\citep{gu2025breakingmodalitybarrieruniversal, yuksekgonul2023when}. Empirical evidence reveals that applying HNM to specific domains often yields suboptimal performance: as illustrated in Fig.~\ref{fig:combined_dataset}, semantic distributions in domain-specific settings are highly compressed, resulting in a substantially higher density of FNs which are hard to identify. To mitigate the cross-modal misleading problem where one-hot labels falsely penalize potential positives, CUSA~\citep{huang2024cross} and SoftCLIP~\citep{gao2024softclip} leverage soft labels derived from unimodal pre-trained models. However, this paradigm inherently creates a mismatch with cross-modal alignment, as unimodal similarity distributions cannot faithfully represent the cross-modal alignment required for TBIR, leading to inconsistent performance gains in domain-specific retrieval.

To address these limitations, we propose the \textbf{Semantic-Aware Fine-Tuning} (SAFT) framework. SAFT extends standard contrastive learning with two complementary components: Semantic-Aware Soft-Label Supervision (SASS), which accommodates potential positive text-image matches via a soft cross-modal alignment distribution, and Intra-modal Structural Distillation (ISD), which preserves relative correlations among images within the visual modality. As shown in Fig~\ref{fig:first}, this integrated framework consistently improves domain-specific retrieval performance.

In summary, the main contributions are as follows:

\begin{itemize}
    \item We introduce SecMM-TBIR, a multi-match benchmark for surveillance scenarios. Built via our DSMM-TBIR data engine, it comprises 50k images and 200 queries across pedestrian and vehicle domains, and will be publicly released to facilitate further research.
    \item We propose SAFT, a systematic fine-tuning framework for CLIP-like~\citep{radford2021learning} models to address performance degradation caused by domain-specific semantic compression. Integrating SASS and ISD, it effectively mitigates false negatives in contrastive learning.
    \item Extensive experiments across various models demonstrate that SAFT achieves an average mAP@20 gain of 7.8 points on SecMM-TBIR, while also boosting general-domain performance. Moreover, SAFT generalizes well to other domains and enhances compositional reasoning.
\end{itemize}

\section{Related Works}

\subsection{Text-Based Image Retrieval Benchmarks}
Rigorous benchmarks lay the foundation for evaluating TBIR systems. Existing testbeds are typically categorized into general-domain and specific-domain. General-domain benchmarks, such as MS-COCO and Flickr30K, encompass an extensive range of scenes, serving as the primary standard for evaluating cross-modal representation alignment. Conversely, domain-specific benchmarks concentrate on constrained vertical scenarios. Within this category, Text-based Person Retrieval (TBPR) represents a well-established task, evaluated on datasets such as CUHK-PEDES~\citep{Li_2017_CVPR}, RSTPReid~\citep{zhu2021dssl}, ICFG-PEDES~\citep{ding2021semantically} and SYNTH-PEDES~\citep{zuo2024pliplanguageimagepretrainingperson}.

% \textcolor{red}{Although traditional benchmarks have driven substantial progress, they are structurally limited by rigid single-match mapping and verbose caption styles. This formulation deviates from real-world, multi-match industrial requirements, where queries are concise and correspond to multiple visual candidates.} 

Although traditional benchmarks have driven substantial progress, they are structurally limited by rigid single-match mappings and verbose captions, diverging from industrial realities that demand task-aligned concise queries and multi-match retrieval. To bridge this gap, several benchmarks have emerged, with InQuire~\citep{vendrow2024inquire} and FSIR-BD~\citep{idan2026few} being the closest to our work. InQuire focuses on expert-level ecological retrieval but demands intensive manual labeling, limiting its scalability to broader scenarios. FSIR-BD, built on the general-purpose Visual Genome dataset~\citep{Krishna2016VisualGC}, relies heavily on manual query annotation and struggles to provide comprehensive semantic coverage required for real-world scenarios. Consequently, balancing annotation scalability, public accessibility, and diverse semantic coverage remains an open challenge in the construction of DSMM-TBIR benchmarks.

% \subsection{Vision-Language Pre-training}

\begin{figure*}[!t]
    \centering
    \includegraphics[width=1.0\linewidth]{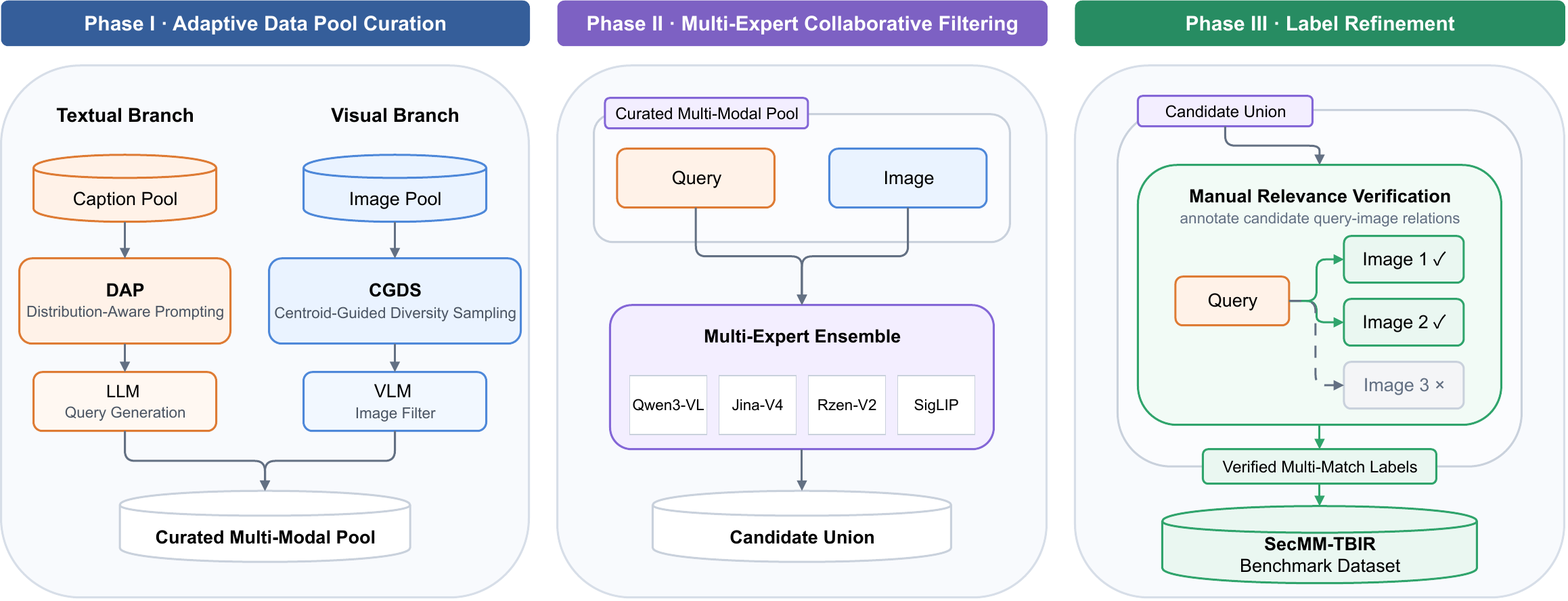}
    \caption{\textbf{DSMM-TBIR Data Engine.} 
    Phase I constructs a curated multi-modal pool through Distribution-Aware Prompting (DAP) and Centroid-Guided Diversity Sampling (CGDS). Phase II employs multiple expert embedders to pre-annotate query-image pairs. Finally, Phase III manually filters these candidates to obtain verified multi-match labels for SecMM-TBIR.}
    \label{fig:DSMM-TBIR Data Engine}
\end{figure*}

\subsection{False Negatives in Contrastive Learning}

Foundational vision-language models, such as CLIP and ALIGN~\citep{jia2021align}, establish cross-modal representation alignment via contrastive pre-training on massive data~\citep{schuhmann2021laion400m, schuhmann2022laion5b} with large batch sizes, which inevitably introduce FNs. Recent universal embedding models~\citep{qwen3vlembedding, jian2025rzenembed} utilize threshold-based HNM to mitigate FNs while capturing HNs. Yet, this yields suboptimal performance in domain-specific settings, as threshold-based methods struggle to disentangle HNs from FNs under concentrated semantic distributions.

To address this, a prevalent alternative leverages soft-label supervision. Within specific domains, e-CLIP~\citep{shin2022clip} introduced catalog-based soft labels to address duplicate products. While MedCLIP~\citep{wang2022medclip} proposed a soft semantic matching loss based on multi-hot labels, it is applicable only to closed-set retrieval. ~\citet{ko2025bringing} leveraged pre-trained models to extract domain-specific features and graph representations for joint supervision. Meanwhile, CellCLIP~\citep{lu2026cellclip} employed DINOv2~\citep{oquab2023dinov2} to construct a continuous visual similarity matrix, adopting the Continuously Weighted Contrastive Loss~\citep{srinivasa2023cwcl} for alignment guidance. Paradigms such as ICSD~\citep{chen2025intra} and CUSA leverage unimodal soft labels to supervise cross-modal alignment. Notably, ICSD suffers from architectural coupling that limits deployment. In contrasct, CUSA remains architecture-independent but underperforms as unimodal statistics fail to reflect true cross-modal matching probabilities. In summary, existing methods rely heavily on rigid domain priors or approximate cross-modal relationships using unimodal statistics, leaving domain-specific TBIR still underexplored.

\section{Methodology}

\begin{figure*}[t]
    \centering
    \begin{subfigure}[b]{0.38\textwidth}
        \centering
        \includegraphics[width=1.0\linewidth]{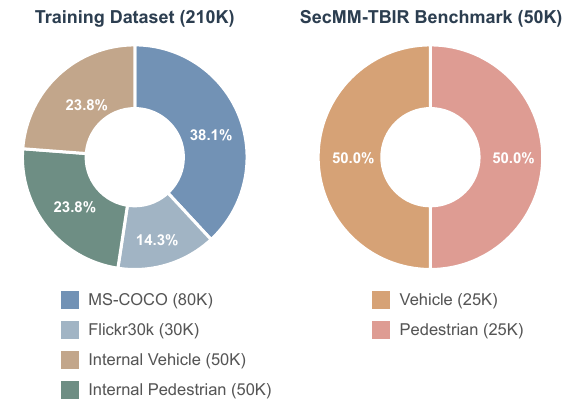}
        % \caption{}
        \label{fig:dataset_train_benchmark}
    \end{subfigure}
    \hfill
    \begin{subfigure}[b]{0.6\textwidth}
        \centering
        \includegraphics[width=1.0\linewidth]{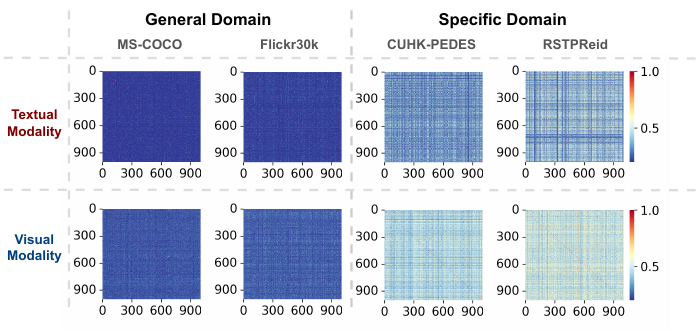}
        % \caption{}
        \label{fig:heatmap}
    \end{subfigure}
    \caption{\textbf{Dataset Statistics and Semantic Compression  Analysis.} Left: Statistics of training set and SecMM-TBIR benchmark. Right: Visualizations of cross-sample similarities (1,000 random samples), where specific domains exhibit higher similarities.}
    \label{fig:combined_dataset}
\end{figure*}

\subsection{DSMM-TBIR Data Engine}
We introduce the DSMM-TBIR Engine, a versatile data pipeline designed for the automated construction of benchmarks across diverse vertical domains. It is designed to overcome four critical limitations in current benchmarks. General-domain benchmarks fail to represent vertical industrial scenarios; we addressed \textit{domain incompatibility} by targeting constrained settings. Unlike traditional datasets that rely on detailed captions, we generate concise queries which are more compatible for practical applications to resolve \textit{caption-query disparity}. We construct a large-scale testbed comprising 50k images and 200 comprehensive queries to address the \textit{evaluation saturation} caused by the limited test scales of existing benchmarks. Evaluating models with single-match labels in specific domains prevents standard metrics from reflecting actual retrieval performance. To eliminate the \textit{metric bias}, we utilize a multi-match paradigm and adopt mAP@K as the primary metrics.

As illustrated in Fig~\ref{fig:DSMM-TBIR Data Engine}, VLMs and LLMs enable efficient adaptation to specific scenarios, allowing rapid TBIR evaluation without extensive manual labeling. DSMM-TBIR data engine comprises three sequential phases:

\paragraph{Phase 1: Adaptive Data Pool Curation}
We utilize both textual and visual branches to establish a representative domain-specific multi-modal data pool.

For the textual branch, we develop \textbf{Distribution-Aware Prompting (DAP)} to generate queries that comprehensively cover the scenarios within the task domain. Specifically, we perform a statistical analysis of the textual distribution in the caption pool and incorporate the resulting statistics into prompt engineering to guide LLM-based query generation. Details of the prompting procedure are provided in the Supplementary Material. This process transforms raw metadata into natural queries that comprehensively reflect the semantic distributions of real-world environments.

For the visual branch, \textbf{Centroid-Guided Diversity Sampling (CGDS)} is designed to construct a diverse image gallery. Under this strategy, the raw image pool $\mathcal{I}$ is projected into a high-dimensional embedding space via a versatile visual representation model~\citep{simeoni2025dinov3}, and partitioned into $\tilde{K}$ distinct semantic regions using K-means clustering. Within each cluster, images are randomly sampled to capture intra-class variance and visual diversity, ensuring balanced and comprehensive coverage of the target domain.

\paragraph{Phase 2: Multi-Expert Collaborative Filtering (MECF)}
Individual VLMs suffer from limited cross-domain generalization, delivering uneven performance across domains, with each model exhibiting varying performance in different scenarios. To alleviate this limitation, we employ an ensemble of four diverse embedders~\citep{qwen3vlembedding, jian2025rzenembedcomprehensivemultimodalretrieval, günther2025jinaembeddingsv4universalembeddingsmultimodal, tschannen2025siglip}.

For each query $q$, we retrieve the top-$\hat{K}$ visual candidates with the highest cosine similarity in the corresponding embedding space. To minimize false negatives, we define the preliminary candidate set $\mathcal{C}(q)$ as the union of the top-$\hat{K}$ retrieval results across all $M$ models:

\begin{equation}
\mathcal{C}(q) =
\bigcup_{i=1}^{M}
\left\{
x \in \mathcal{I}
\;\middle|\;
\operatorname{rank}_{\mathcal{I}}
\left(
\operatorname{sim}(\phi_i(q), \psi_i(x))
\right)
\leq \hat{K}
\right\}.
\end{equation}

where $q$ denotes the query,  $x$ is a candidate image, $\phi_i$ and $\psi_i$ are the text and image embedders of the $i$-th model, and $\operatorname{sim}(\cdot)$ denotes cosine similarity.

\paragraph{Phase 3: Label Refinement}
Annotators manually inspect the visual candidates in $\mathcal{C}(q)$ and verify their relevance to the query $q$, filtering out residual false positives to obtain the final multi-match labels.

\subsection{Rethinking Fine-Tuning Paradigms}
\label{sec:rethinking_methods}
To characterize the semantic density of TBIR datasets, we randomly sample 1,000 instances from each dataset and compute pairwise cosine similarities in the embedding space. As shown in Fig~\ref{fig:combined_dataset}, specific domains exhibit substantially more compressed latent spaces. Consequently, semantically related yet unpaired samples are more prevalent within a mini-batch and be treated as negatives under standard image-text contrastive (ITC) learning. This motivates us to revisit three common fine-tuning choices in TBIR tasks:

\textbf{Text Encoder Joint Fine-Tuning} in specific domains often induces overfitting to the textual patterns of training sets.

\textbf{Standard Image Self-Supervised (ISS) Objectives} force the separation of unpaired images in a mini-batch which are semantically near-identical, yielding distorted rather than discriminative representations.

\textbf{Hard Negative Mining} defines a threshold $\alpha$ by shifting the similarity between the query embedding $e_q$ and its paired positive target embedding $e_t^{+}$ with a margin parameter $\beta$:

\begin{equation}
\alpha = \cos(e_q, e_t^{+}) + \beta
\end{equation}

Within a batch of candidate negatives $\{e_t^{-}\}$, any sample satisfying $\cos(e_q, e_t^{-}) > \alpha$ is excluded as a false negative, while the remaining instances are ranked by similarity to select the top-$K$ hard negatives. However, this threshold-based mechanism enforces a rigid separation between FNs and HNs, failing to reliably distinguish them in semantically compressed training sets.

\begin{figure*}[htbp]
    \centering
    \includegraphics[width=1.0\linewidth]{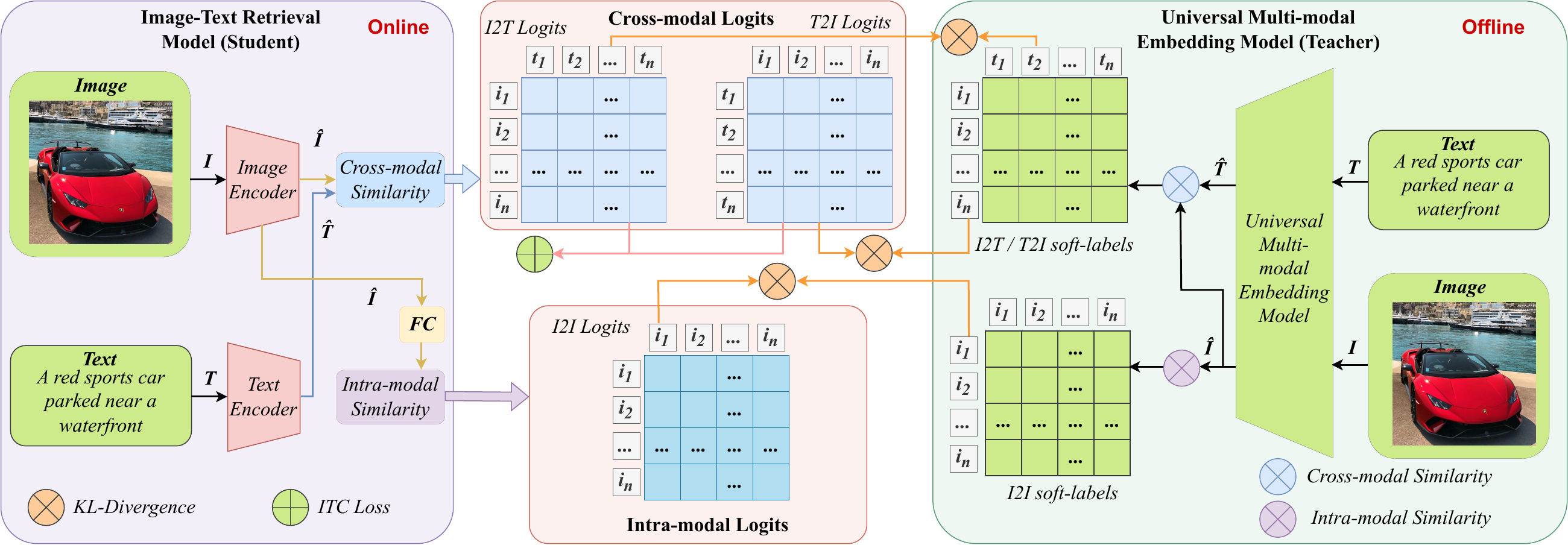}
    \caption{\textbf{Overview of SAFT Framework.} A frozen universal multi-modal embedding model serves as the teacher, providing cross-modal soft targets for SASS and image-to-image structural targets for ISD, complementing the standard ITC objective.}
    \label{fig:SAFT_main}
\end{figure*}

\subsection{Loss Functions}
The standard ITC loss in CLIP pre-training is formulated as:
\begin{equation}
\label{eq:itc_loss}
\mathcal{L}_{\mathrm{ITC}} = -\frac{1}{2N} \sum_{i=1}^{N} \sum_{j=1}^{N} \big( q_{i,j} \log p_{i,j} + q_{j,i} \log p_{j,i} \big)
\end{equation}
where $N$ represents batch size, $q$ represents rigid one-hot labels. The predicted image-to-text (I2T) probability $p_{i,j}$ measures the probability of matching the $i$-th image to the $j$-th text, while the text-to-image (T2I) probability $p_{j,i}$ measures the probability of matching the $i$-th text to the $j$-th image:

\begin{equation}
\label{eq:p_prob}
\begin{aligned}
p_{i,j} &= \frac{\exp( (e^{\mathrm{v}}_i)^\top e^{\mathrm{t}}_j / \tau)}{\sum_{k=1}^{N} \exp( (e^{\mathrm{v}}_i)^\top e^{\mathrm{t}}_k / \tau)}, \\
p_{j,i} &= \frac{\exp( (e^{\mathrm{t}}_i)^\top e^{\mathrm{v}}_j / \tau)}{\sum_{k=1}^{N} \exp( (e^{\mathrm{t}}_i)^\top e^{\mathrm{v}}_k / \tau)}
\end{aligned}
\end{equation}
where $e^{\mathrm{v}}_i$ and $e^{\mathrm{t}}_j$ denote the normalized visual and textual embeddings for the $i$-th and $j$-th instances within a mini-batch respectively, and $\tau$ represents the temperature.

% \textcolor{blue}{To exploit the continuous semantic relationships between non-paired samples, we utilize the \textbf{SASS} to shift the training paradigm from rigid one-hot classification to cross-modal distributional alignment.} 
We employ UniME-V2~\citep{unimev2}, a universal cross-modal embedder to generate reference visual embeddings $E^{\mathrm{v}}$ and textual embeddings $E^{\mathrm{t}}$, which preserve deep semantic inter-dependencies. For a given training batch, these embeddings are used to derive soft-label over image-text pairs:
\begin{equation}
\label{eq:q_soft}
\begin{aligned}
r_{i,j} &= \frac{\exp( (E^{\mathrm{v}}_i)^\top E^{\mathrm{t}}_j / \tau_{o})}{\sum_{k=1}^{N} \exp( (E^{\mathrm{v}}_i)^\top E^{\mathrm{t}}_k / \tau_{o})}, \\
r_{j,i} &= \frac{\exp( (E^{\mathrm{t}}_i)^\top E^{\mathrm{v}}_j / \tau_{o})}{\sum_{k=1}^{N} \exp( (E^{\mathrm{t}}_i)^\top E^{\mathrm{v}}_k / \tau_{o})}
\end{aligned}
\end{equation}
Let $S_{\mathrm{i2t}} = [p_{i,j}]$ and $S_{\mathrm{t2i}} = [p_{j,i}]$ denote the predicted cross-modal distributions of the lightweight student model. Similarly, $T_{\mathrm{i2t}} = [r_{i,j}]$ and $T_{\mathrm{t2i}} = [r_{j,i}]$ denote the target distributions generated by the teacher embedder. All distributions are in $\mathbb{R}^{N \times N}$. The SASS loss is formulated as a bidirectional Kullback-Leibler (KL) divergence to encourage robust cross-modal alignment:

\begin{equation}
\label{eq:l_sass}
\begin{aligned}
\mathcal{L}_{\mathrm{SASS}} = \frac{1}{4} \Big( & \mathcal{D}_{\mathrm{KL}}(T_{\mathrm{i2t}} \parallel S_{\mathrm{i2t}}) + \mathcal{D}_{\mathrm{KL}}(T_{\mathrm{t2i}} \parallel S_{\mathrm{t2i}}) \\
+ & \mathcal{D}_{\mathrm{KL}}(S_{\mathrm{i2t}} \parallel T_{\mathrm{i2t}}) + \mathcal{D}_{\mathrm{KL}}(S_{\mathrm{t2i}} \parallel T_{\mathrm{t2i}}) \Big)
\end{aligned}
\end{equation}

While SASS improves cross-modal alignment, it does not explicitly preserve the relational structure within the visual modality. We therefore propose \textbf{Intra-modal Structural Distillation (ISD)}, which transfers the teacher's fine-grained image-image relationships to the student. Specifically, for each mini-batch, we use the teacher's visual embeddings $E^{\mathrm{v}}$ to construct an intra-modal soft targets:

\begin{equation}
\label{eq:r_v2v}
r^{\mathrm{v}}_{i,j} = \frac{\exp( (E^{\mathrm{v}}_i)^\top E^{\mathrm{v}}_j / \tau_{o})}{\sum_{k=1}^{N} \exp( (E^{\mathrm{v}}_i)^\top E^{\mathrm{v}}_k / \tau_{o})}
\end{equation}
We define the teacher's intra-modal structural distribution as $T_{\mathrm{v2v}} = [r^{\mathrm{v}}_{i,j}] \in \mathbb{R}^{N \times N}$, the student's predicted distribution is denoted as $S_{\mathrm{v2v}} = [p^{\mathrm{v}}_{i,j}] \in \mathbb{R}^{N \times N}$, where $p^{\mathrm{v}}_{i,j}$ is computed based on the visual embeddings.
\begin{equation}
\label{eq:p_v2v}
p^{\mathrm{v}}_{i,j} = \frac{\exp( (e^{\mathrm{v}}_i)^\top e^{\mathrm{v}}_j / \tau)}{\sum_{k=1}^{N} \exp( (e^{\mathrm{v}}_i)^\top e^{\mathrm{v}}_k / \tau)}
\end{equation}
The ISD loss is then formulated as the KL divergence between these two intra-modal distributions:
\begin{equation}
\label{eq:l_isd}
\mathcal{L}_{\mathrm{ISD}} = \mathcal{D}_{\mathrm{KL}}(T_{\mathrm{v2v}} \parallel S_{\mathrm{v2v}})
\end{equation}
Finally, the joint training objective is formulated as:
\begin{equation}
\label{eq:l_total}
\mathcal{L}_{\mathrm{SAFT}} = \mathcal{L}_{\mathrm{ITC}} + \alpha \cdot \mathcal{L}_{\mathrm{SASS}} + \beta \cdot \mathcal{L}_{\mathrm{ISD}} 
\end{equation}
where $\alpha$ and $\beta$ are balancing hyperparameters.

\noindent \textbf{Difference from Unimodal Proxy Supervision.} Unlike prior works that use unimodal soft labels as proxies for cross-modal supervision, SAFT leverages cross-modal similarity distributions derived from a universal multi-modal embedder, enabling more faithful cross-modal supervision.

\begin{table}[!b]
\centering
\footnotesize
\setlength{\tabcolsep}{3pt}

\begin{tabular*}{\columnwidth}{@{\extracolsep{\fill}} l *{6}{c} }
    \toprule
    \multirow{3}{*}{\textbf{Model}} & \multicolumn{6}{c}{\textbf{SecMM-TBIR}} \\
    \cmidrule(lr){2-7}
    & \multicolumn{3}{c}{\textbf{Pedestrian}} & \multicolumn{3}{c}{\textbf{Vehicle}} \\
    \cmidrule(lr){2-4} \cmidrule(lr){5-7} 
    & m@10 & m@20 & m@30 & m@10 & m@20 & m@30 \\
    \midrule
    
    % --- TinyCLIP-32M Group ---
    TinyCLIP-22M/32 & 32.5 & 25.1 & 23.3 & 45.4 & 38.0 & 33.5 \\
    \quad + ITC                   & 38.0 & 29.8 & 27.8 & 51.7 & 42.5 & 37.0 \\
    \quad + CUSA                  & 36.2 & 29.3 & 27.6 & 48.7 & 40.7 & 34.8 \\
    \quad \textbf{+ SAFT (ours)}  & \textbf{40.7} & \textbf{32.1} & \textbf{29.6} & \textbf{60.1} & \textbf{51.3} & \textbf{45.8} \\
    \cmidrule(lr){1-7}
    
    % --- TinyCLIP-63M Group ---
    TinyCLIP-45M/32 & 40.0 & 32.7 & 30.4 & 53.3 & 45.2 & 40.2 \\
    \quad + ITC                   & 45.4 & 37.6 & 35.3 & 67.6 & 56.7 & 50.7 \\
    \quad + CUSA                  & 42.3 & 34.5 & 32.4 & 60.2 & 51.1 & 46.6 \\
    \quad \textbf{+ SAFT (ours)}  & \textbf{50.5} & \textbf{41.5} & \textbf{38.4} & \textbf{73.3} & \textbf{64.6} & \textbf{58.5} \\
    \cmidrule(lr){1-7}
    
    % --- MobileCLIP-S0 Group ---
    MobileCLIP-S0 & 49.7 & 41.0 & 38.5 & 60.1 & 52.2 & 46.5 \\
    \quad + ITC                   & 53.1 & 44.2 & 41.3 & 65.8 & 57.4 & 51.6 \\
    \quad + CUSA                  & 53.7 & 45.8 & 42.7 & 69.2 & 59.4 & 53.3 \\
    \quad \textbf{+ SAFT (ours)}  & \textbf{59.8} & \textbf{49.7} & \textbf{46.7} & \textbf{77.7} & \textbf{67.9} & \textbf{61.7} \\
    \cmidrule(lr){1-7}
    
    % --- MobileCLIP-S1 Group ---
    MobileCLIP-S1 & 52.1 & 42.6 & 39.8 & 65.2 & 57.2 & 51.7 \\
    \quad + ITC                   & 58.0 & 48.4 & 44.9 & 64.8 & 59.4 & 54.8 \\
    \quad + CUSA                  & 59.6 & 49.1 & 46.3 & 69.0 & 62.2 & 56.8 \\
    \quad \textbf{+ SAFT (ours)}  & \textbf{65.0} & \textbf{54.9} & \textbf{52.1} & \textbf{79.2} & \textbf{73.0} & \textbf{66.8} \\
    \cmidrule(lr){1-7}
    
    % --- OpenCLIP-B/32 Group ---
    OpenCLIP-B/32 & 34.0 & 27.4 & 25.5 & 55.6 & 46.5 & 41.6 \\
    \quad + ITC                   & 41.0 & 33.2 & 31.4 & 63.5 & 55.6 & 49.3 \\
    \quad + CUSA                  & 39.2 & 31.6 & 29.3 & 63.0 & 55.6 & 49.7 \\
    \quad \textbf{+ SAFT (ours)}  & \textbf{47.5} & \textbf{39.0} & \textbf{36.3} & \textbf{73.1} & \textbf{64.8} & \textbf{58.5} \\
    \cmidrule(lr){1-7}
    
    % --- OpenCLIP-B/16 Group ---
    OpenCLIP-B/16 & 40.5 & 32.9 & 30.4 & 56.4 & 48.0 & 42.9 \\
    \quad + ITC                   & 47.8 & 38.8 & 36.3 & 65.4 & 56.8 & 51.6 \\
    \quad + CUSA                  & 49.3 & 40.1 & 37.6 & 65.8 & 57.0 & 52.2 \\
    \quad \textbf{+ SAFT (ours)}  & \textbf{57.9} & \textbf{46.9} & \textbf{44.0} & \textbf{75.4} & \textbf{68.7} & \textbf{62.8} \\
    \bottomrule
\end{tabular*}
\caption{\textbf{Main Results on SecMM-TBIR Benchmark.}}
\label{tab:results_SecMMTBIR}
\end{table}

\begin{table*}[!t]
    \centering
    \footnotesize
    \setlength{\tabcolsep}{0pt}
    \begin{tabular*}{\textwidth}{@{\extracolsep{\fill}} cc llllll llllll @{}}
    \toprule
    \multicolumn{2}{c}{\textbf{Fine-tuned Encoders}} & \multicolumn{6}{c}{\textbf{Pedestrian (mAP@20 $\uparrow$)}} & \multicolumn{6}{c}{\textbf{Vehicle (mAP@20 $\uparrow$)}} \\
    \cmidrule(lr){1-2} \cmidrule(lr){3-8} \cmidrule(lr){9-14}
    Image & Text & T-22 & T-45 & M-S0 & M-S1 & B/16 & B/32 & T-22 & T-45 & M-S0 & M-S1 & B/16 & B/32 \\
    \midrule
    % 原始基线
    $\text{--}$ & $\text{--}$ & 25.1 & 32.7 & 41.0 & 42.6 & 32.9 & 27.4 & 38.0 & 45.2 & 52.2 & 57.2 & 48.0 & 46.5 \\
    $\checkmark$ & $\text{--}$ & $\mathbf{29.8}$ & $\mathbf{37.6}$ & $\mathbf{44.2}$ & $\mathbf{48.4}$ & $\mathbf{38.8}$ & $\mathbf{33.2}$ & $\mathbf{42.5}$ & $\mathbf{56.7}$ & $\mathbf{57.9}$ & $\mathbf{61.4}$ & $\mathbf{56.8}$ & $\mathbf{57.3}$ \\
    $\checkmark$ & $\checkmark$ & 
    $26.5^{\textcolor{red}{-3.3}}$ & 
    $35.8^{\textcolor{red}{-1.8}}$ & 
    $40.0^{\textcolor{red}{-4.2}}$ & 
    $43.5^{\textcolor{red}{-4.9}}$ & 
    $38.1^{\textcolor{red}{-0.7}}$ & 
    $29.2^{\textcolor{red}{-4.0}}$ & 
    $40.0^{\textcolor{red}{-2.5}}$ & 
    $54.5^{\textcolor{red}{-2.2}}$ & 
    $55.9^{\textcolor{red}{-2.0}}$ & 
    $59.7^{\textcolor{red}{-1.7}}$ & 
    $56.1^{\textcolor{red}{-0.7}}$ & 
    $55.1^{\textcolor{red}{-2.2}}$ \\
    \bottomrule
    \end{tabular*}
    \caption{\textbf{Ablation Study on Encoder Optimization.} Results obtained by fine-tuning different encoder modules are reported.}
    \label{tab:encoder_ablation}
\end{table*}

\begin{table}[!t]
\centering
\footnotesize
\setlength{\tabcolsep}{3pt}

\begin{tabular*}{\columnwidth}{@{\extracolsep{\fill}} l *{6}{c} }
    \toprule
    \multirow{3}{*}{\textbf{Model}} & \multicolumn{6}{c}{\textbf{General Benchmarks}} \\
    \cmidrule(lr){2-7}
    & \multicolumn{3}{c}{\textbf{Flickr30K}} & \multicolumn{3}{c}{\textbf{MS-COCO}} \\
    \cmidrule(lr){2-4} \cmidrule(lr){5-7} 
    & R@1 & R@5 & R@10 & R@1 & R@5 & R@10 \\
    \midrule
    
    % --- TinyCLIP-32M Group ---
    TinyCLIP-22M/32                       & 51.8 & 77.2 & 85.3 & 28.4 & 53.0 & 64.5 \\
    \quad + ITC                        & 56.1 & 80.6 & 88.4 & 31.6 & 58.0 & 69.3 \\
    \quad + CUSA                       & 56.9 & 81.7 & 88.4 & 33.5 & 61.0 & 72.2 \\
    \quad \textbf{+ SAFT (ours)}       & \textbf{57.3} & \textbf{82.9} &  \textbf{88.7} &  \textbf{35.0} &  \textbf{62.6} &  \textbf{73.7} \\
    \cmidrule(lr){1-7}
    
    % --- TinyCLIP-63M Group ---
    TinyCLIP-45M/32                  & 64.0 & 86.0 & 92.1 & 36.9 & 62.6 & 73.1\\
    \quad + ITC                   & 67.6 & 89.1 & 93.8 & 41.1 & 67.7 & 78.0 \\
    \quad + CUSA                  & 68.4 & 89.6 & 94.5 & 42.2 & 69.5 & 79.7 \\
    \quad \textbf{+ SAFT (ours)}  & \textbf{70.0} & \textbf{89.9} & \textbf{94.8} & \textbf{43.5} & \textbf{70.9} & \textbf{80.8} \\
    \cmidrule(lr){1-7}
    
    % --- MobileCLIP-S0 Group ---
    MobileCLIP-S0                 & 66.8 & 88.7 & 93.1 & 39.6 & 65.9 & 75.6\\
    \quad + ITC                   & 70.8 & 90.9 & 93.6 & 44.6 & 72.5 & 81.6 \\
    \quad + CUSA                  & 71.3 & 91.7 & 95.3 & 45.6 & 73.2 & 82.8 \\
    \quad \textbf{+ SAFT (ours)}  & \textbf{71.9} & \textbf{91.8} & \textbf{95.5} & \textbf{46.3} & \textbf{73.3} & \textbf{83.0} \\
    \cmidrule(lr){1-7}
    
    % --- MobileCLIP-S1 Group ---
    MobileCLIP-S1                 & 71.4 & 90.8 & 95.0 & 43.7 & 69.1 & 78.2\\
    \quad + ITC                   & 75.9 & 93.1 & 94.9 & 49.2 & 75.4 & 83.6 \\
    \quad + CUSA                  & 76.8 & 93.7 & 96.7 & 50.0 & 76.0 & 84.3 \\
    \quad \textbf{+ SAFT (ours)}  & \textbf{77.4} & \textbf{94.2} & \textbf{96.9} & \textbf{50.7} & \textbf{77.0} & \textbf{85.4} \\
    \cmidrule(lr){1-7}
    
    % --- OpenCLIP-B/32 Group ---
    OpenCLIP-B/32                 & 61.1 & 85.0 & 90.9 & 37.1 & 62.3 & 72.7\\
    \quad + ITC                   & 66.3 & 88.6 & 93.1 & 42.8 & 71.2 & 81.1 \\
    \quad + CUSA                  & \textbf{69.3} & 90.2 & 94.5 & 43.7 & 71.2 & 80.7 \\
    \quad \textbf{+ SAFT (ours)}  & \textbf{69.3} & \textbf{90.5} & \textbf{94.7} & \textbf{45.0} & \textbf{72.1} & \textbf{81.6} \\
    \cmidrule(lr){1-7}
    
    % --- OpenCLIP-B/16 Group ---
    OpenCLIP-B/16                 & 67.5 & 88.4 & 93.0 & 40.1 & 66.0 & 75.6 \\
    \quad + ITC                   & 73.9 & 92.2 & 95.5 & 47.1 & 73.3 & 82.5 \\
    \quad + CUSA                  & 75.7 & 93.1 & 96.3 & 48.1 & 74.6 & 83.5 \\
    \quad \textbf{+ SAFT (ours)}  & \textbf{76.1} & \textbf{93.3} & \textbf{96.5} & \textbf{49.3} & \textbf{75.7} & \textbf{84.5} \\
    \bottomrule
\end{tabular*}
\caption{\textbf{Main Results on General Benchmarks.}}
\label{tab:general_domain_test}
\end{table}

\section{Experiments}
\subsection{Experiment Setup}
\label{sec:implementation}

\paragraph{Evaluation Benchmarks.} As illustrated in Fig~\ref{fig:combined_dataset}, we perform our evaluation on SecMM-TBIR, which comprises 50k surveillance images and 200 comprehensive queries across two subdomains: pedestrian and vehicle. More details are provided in the Supplementary Material.

\paragraph{Training Sets.} The training sets comprise two standard general datasets, augmented by a specialized internal dataset which are decoupled from SecMM-TBIR benchmark:
\begin{itemize}
    \item \textbf{General Multi-modal Dataset}: Comprising Flickr30K and MS-COCO training splits to ensure data balance, this dataset is explicitly utilized to preserve robust pretrained general cross-modal alignment.
    \item \textbf{Specialized Internal Dataset}: Consisting of real-world image-text pairs from surveillance scenarios, the dataset focuses on pedestrian and vehicle domains to enhance discriminability. Specifically, it's built via Rex-Omni~\citep{jiang2025detectpointprediction} for video frame object detection and Qwen3-VL~\cite{Qwen3-VL} together with Qwen-3.5~\citep{qwen3_5} for high-fidelity captions.
\end{itemize}

\paragraph{Baseline Configurations.} To demonstrate broad applicability, we select a broad spectrum of CLIP-like models: TinyCLIP~\citep{tinyclip}, MobileCLIP~\citep{faghri2025mobileclip2}, OpenCLIP~\citep{cherti2023reproducible}. As our first baseline, we freeze the text encoder and optimize only the visual branch via the standard ITC objective. Our second comparative baseline CUSA establishes cross-modal alignment via teacher-derived unimodal priors without introducing additional architectural components, and we upgrade its teacher to the stronger UniME-V2 to ensure fairness.

\paragraph{Implementation Details.} All experiments are conducted on a single NVIDIA RTX 5090 GPU (32GB VRAM). The model is fine-tuned for 25k iterations via AdamW~\citep{loshchilov2018adamw} with batch size of 128. Specifically, the learning rate linearly warms up from $5 \times 10^{-7}$ to $5 \times 10^{-6}$ over the first $10\%$ of steps, then follows cosine annealing with a weight decay of $0.05$. Further details are provided in the Supplementary Material.

\begin{table}[!b]
    \centering
    \footnotesize
    \setlength{\tabcolsep}{3.5pt}
    \begin{tabular}{lcccc}
        \toprule
        \multirow{2}{*}{\textbf{Model}} & \multicolumn{2}{c}{\textbf{General (R@1 $\uparrow$)}} & \multicolumn{2}{c}{\textbf{SecMM-TBIR (m@20 $\uparrow$)}} \\
        \cmidrule(lr){2-3} \cmidrule(lr){4-5} 
         & \textbf{Flickr} & \textbf{COCO} & \textbf{Pedestrian} & \textbf{Vehicle} \\
        \midrule
        Tiny-22M/32 & 51.8 & 28.4 & 25.1 & 38.0 \\
        Tiny-45M/32 & 64.0 & 36.9 & 32.7 & 45.2 \\
        MobileCLIP-S0 & 66.8 & 39.6 & 41.0 & 52.2 \\
        MobileCLIP-S1 & 71.4 & 43.7 & 42.6 & 57.2 \\
        OpenCLIP-B/32 & 61.1 & 37.1 & 27.4 & 46.5 \\
        OpenCLIP-B/16 & 67.5 & 40.1 & 32.9 & 48.0 \\
        Qwen3-VL-2B & \textbf{90.2} & \textbf{77.3} & 45.1 & 64.3 \\
        UniMEV2-2B & 89.9 & 65.1 & \textbf{57.4} & \textbf{68.9} \\
        \bottomrule
    \end{tabular}
    \caption{\textbf{Zero-shot Generalization of Foundation Models.}}
    \label{tab:foundation_model_generalization_gap}
\end{table}

\begin{table*}[htbp]
    \centering
    \small
    \footnotesize
    \setlength{\tabcolsep}{0pt} 
    \begin{tabular*}{\textwidth}{@{\extracolsep{\fill}} l c ccc ccc ccc ccc ccc ccc @{}}
    \toprule
    \multirow{3}{*}{\textbf{Backbone}} & \multirow{3}{*}{\textbf{Baseline}} & \multicolumn{9}{c}{\textbf{Pedestrian Dataset}} & \multicolumn{9}{c}{\textbf{Vehicle Dataset}} \\
    \cmidrule(lr){3-11} \cmidrule(lr){12-20}    & & \multicolumn{3}{c}{$K = 32$} & \multicolumn{3}{c}{$K = 64$} & \multicolumn{3}{c}{$K = 96$} & \multicolumn{3}{c}{$K = 32$} & \multicolumn{3}{c}{$K = 64$} & \multicolumn{3}{c}{$K = 96$} \\
    \cmidrule(lr){3-5} \cmidrule(lr){6-8} \cmidrule(lr){9-11} \cmidrule(lr){12-14} \cmidrule(lr){15-17} \cmidrule(lr){18-20}
    & (w/o) & $\beta$=--.1 & $\beta$=.0 & $\beta$=.1 & $\beta$=--.1 & $\beta$=.0 & $\beta$=.1 & $\beta$=--.1 & $\beta$=.0 & $\beta$=.1 & $\beta$=--.1 & $\beta$=.0 & $\beta$=.1 & $\beta$=--.1 & $\beta$=.0 & $\beta$=.1 & $\beta$=--.1 & $\beta$=.0 & $\beta$=.1 \\
    \midrule
    TinyCLIP-22M/32   & 29.8 / 42.5  & 30.0 & 28.4 & 30.2 & 29.6 & 27.6 & 30.2 & 30.1 & 29.9 & \textbf{30.3} & 42.2 & 40.0 & 42.4 & \textbf{42.7} & 39.3 & 42.4 & 42.5 & 42.5 & 42.4 \\ 
    TinyCLIP-45M/32   & 37.6 / 56.7 & 35.1 & 35.0 & 37.8 & 34.5 & 34.9 & 37.9 & 37.5 & \textbf{38.0} & 37.8 & 51.9 & 50.8 & 55.1 & 53.0 & 49.9 & 55.1 & \textbf{55.8} & 55.7 & 55.6 \\
    MobileCLIP-S0  & 44.2 / 57.9 & 42.6 & 44.4 & 44.3 & 42.7 & \textbf{44.5} & 44.2 & 43.9 & 43.5 & 43.9 & 48.8 & 56.2 & 57.4 & 50.6 & 56.0 & 57.6 & 57.5 & 57.8 & \textbf{57.9} \\
    MobileCLIP-S1  & 48.4 / 61.4 & 47.8 & 48.4 & 48.2 & 47.2 & \textbf{48.6} & 48.2 & 48.0 & 48.0 & 48.1 & 55.6 & 59.4 & 60.4 & 55.0 & \textbf{60.6} & 59.8 & 60.1 & 60.0 & 59.8 \\
    OpenCLIP-B/32  & 33.2 / 57.3 & 30.8 & 33.0 & 33.2 & \textbf{33.3} & 32.8 & 33.2 & 32.9 & 32.8 & 32.7 & 53.6 & 51.7 & \textbf{57.5} & 56.0 & 52.7 & 57.4 & 56.2 & 56.2 & 56.3 \\
    OpenCLIP-B/16  & 38.8 / 56.8 & 37.9 & 38.1 & 38.5 & 38.2 & 38.8 & 38.6 & \textbf{39.0} & 38.7 & 38.9 & 56.3 & 56.2 & 56.6 & 56.8 & 56.8 & 57.0 & 56.8 & \textbf{57.1} & 57.0 \\
    
    \bottomrule
    \end{tabular*}
    \caption{\textbf{Hyperparameter Sensitivity of HNM.} We report mAP@20 (\%) across varying margins $\beta$ and candidate sizes $K$. The \textbf{w/o} column denotes the baseline on pedestrian and vehicle without HNM, illustrating the marginal gains of heuristic tuning.}
    \label{tab:ablation_hnm}
\end{table*}

\begin{table*}[t]
    \centering
    \footnotesize
    \setlength{\tabcolsep}{2pt}
    \begin{tabular*}{\textwidth}{@{\extracolsep{\fill}} ccc cccccc cccccc @{}}
    \toprule
    \multicolumn{3}{c}{\textbf{Loss Term}} 
    & \multicolumn{6}{c}{\textbf{Pedestrian (mAP@20 $\uparrow$)}} 
    & \multicolumn{6}{c}{\textbf{Vehicle (mAP@20 $\uparrow$)}} \\
    \cmidrule(lr){1-3} \cmidrule(lr){4-9} \cmidrule(lr){10-15}
    ITC & SASS & ISD & T-22 & T-45 & M-S0 & M-S1 & B/16 & B/32 & T-22 & T-45 & M-S0 & M-S1 & B/16 & B/32 \\
    \midrule
    $\text{--}$ & $\text{--}$ & $\text{--}$ & 25.1 & 32.7 & 41.0 & 42.6 & 32.9 & 27.4 & 38.0 & 45.2 & 52.2 & 57.2 & 48.0 & 46.5 \\

    $\checkmark$ & $\text{--}$ & $\text{--}$ & 29.8 & 37.6 & 44.2 & 48.4 & 38.8 & 33.2 & 42.5 & 56.7 & 57.9 & 61.4 & 56.8 & 57.3 \\

    $\checkmark$ & $\checkmark$ & $\text{--}$ & 31.1 & 40.2 & 48.6 & 54.7 & 45.6 & 37.0 & 50.3 & 64.4 & $66.5$ & $ 72.1$ & 67.5 & 64.7 \\

    $\checkmark$ & $\checkmark$ & $\checkmark$ & $\mathbf{32.1}$ & $\mathbf{41.5}$ & $\mathbf{49.7}$ & $\mathbf{55.0}$ & $\mathbf{46.9}$ & $\mathbf{39.0}$ & $\mathbf{51.3}$ & $\mathbf{64.6}$ & $\mathbf{67.9}$ & $\mathbf{73.0}$ & $\mathbf{68.7}$ & $\mathbf{64.8}$ \\

    \bottomrule
    \end{tabular*}
    \caption{\textbf{Ablation Study on Loss Functions.} We report ITC baseline and incremental gains from adding SASS and ISD across CLIP-like models. SASS consistently improves retrieval performance, and ISD further boosts gains when combined with SASS.}
    \label{tab:ablation_reordered}
\end{table*}
% We report the baseline (ITC) and the effects of SASS and ISD.

\subsection{Main Results}
\label{sec:results}

\paragraph{Results on TBIR Benchmark.}

As summarized in Table~\ref{tab:results_SecMMTBIR}, SAFT consistently outperforms both standard ITC and CUSA across all settings. The selected architectures span transformer and convolution-heavy variants, indicating that the gains of SAFT are not tied to a specific model design. Compared with ITC, SAFT improves average mAP@20 by 5.4 and 10.3 points on pedestrian and vehicle subsets, respectively. SAFT also surpasses CUSA by 5.6 points on pedestrian and 10.7 points on vehicle retrieval, demonstrating the advantage of direct cross-modal distribution supervision over unimodal proxy signals in semantically compressed domains.

\paragraph{Results on General Benchmarks.}
As reported in Table~\ref{tab:general_domain_test}, we report TBIR performance on Flickr30K and MS-COCO. On these general benchmarks, SAFT consistently improves over standard ITC fine-tuning across all reported metrics. It also outperforms CUSA in most settings. While CUSA yields broadly positive improvements over ITC, the gains are less stable across models and task domains.

% MS-COCO and Flickr30K are constructed via a single-match protocol, inherently placing a high premium on model discriminative capability. As reported in Table~\ref{tab:general_domain_test}, while our approach utilizes soft labels, it does not compromise the discriminative power required for general cross-modal retrieval.

\paragraph{Extended Evaluations.} Beyond the main results, we further validate SAFT's general applicability on Fashion200K~\citep{han2017automaticspatiallyawarefashionconcept} and its compositional reasoning ability on ARO~\citep{yuksekgonul2023visionlanguagemodelsbehavelike}. On Fashion200K, SAFT outperforms competing baselines with marked retrieval gains. On ARO, it improves reasoning on both Relation and Attribution subsets. Detailed results are provided in the Supplementary Material.

% Beyond main results, we validate SAFT on Fashion200K~\citep{han2017automaticspatiallyawarefashionconcept} and ARO~\citep{yuksekgonul2023visionlanguagemodelsbehavelike} benchmark. SAFT outperforms competing baselines in the fashion domain and yields stronger compositional reasoning capability. Detailed results are in the Supplementary Material.

\begin{table}[!b]
    \centering
    \footnotesize
    \setlength{\tabcolsep}{6.0pt}
    \begin{tabular}{lcccl}
        \toprule
        \multirow{2}{*}{\textbf{Model}}
        & \multicolumn{2}{c}{\textbf{Pedestrian}}
        & \multicolumn{2}{c}{\textbf{Vehicle}} \\
        \cmidrule(lr){2-3} \cmidrule(lr){4-5}
        & ITC & ITC+ISS & ITC & ITC+ISS \\
        \midrule
        TinyCLIP-22M/32 & 29.8 & $27.2^{\textcolor{red}{-2.6}}$ & 42.5 & $40.9^{\textcolor{red}{-1.6}}$ \\
        TinyCLIP-45M/32   & 37.6 & $33.9^{\textcolor{red}{-3.7}}$ & 56.7 & $53.6^{\textcolor{red}{-3.1}}$ \\
        MobileCLIP-S0  & 44.2 & $44.0^{\textcolor{red}{-0.2}}$ & 57.9 & $56.3^{\textcolor{red}{-1.6}}$ \\
        MobileCLIP-S1  & 48.4 & $48.0^{\textcolor{red}{-0.4}}$ & 61.4 & $59.8^{\textcolor{red}{-1.6}}$ \\
        OpenCLIP-B/32       & 33.2 & $32.0^{\textcolor{red}{-1.2}}$ & 57.3 & $57.3^{0.0}$ \\
        OpenCLIP-B/16       & 38.8 & $38.3^{\textcolor{red}{-0.5}}$ & 56.8 & $56.1^{\textcolor{red}{-0.7}}$ \\
        
        \bottomrule
    \end{tabular}
    \caption{\textbf{Ablation Study on Image Self-Supervision (ISS), evaluated by mAP@20.}}
    \label{tab:issablation}
\end{table}

\subsection{Ablation Studies}

We conduct ablation experiments to analyze the contributions of our framework. To verify the general applicability of our findings, we evaluate these variants across mainstream models optimized with standard ITC objective, on pedestrian and vehicle retrieval sub-tasks.

\paragraph{Foundation Model Generalization Gap.} We first evaluate the zero-shot performance of pre-trained models. As shown in Table~\ref{tab:foundation_model_generalization_gap}, models demonstrating strong performance on general-domain retrieval benchmarks can fail to maintain effectiveness on SecMM-TBIR, underscoring the necessity of benchmarking and fine-tuning in specific domains.

\paragraph{Impact of Text Encoder Tuning.} While joint cross-modal fine-tuning is conventionally assumed to increase alignment, we find it suboptimal across all models and sub-tasks. As reported in Table~\ref{tab:encoder_ablation}, tuning the text encoder consistently degrades performance. We attribute this to the distortion of pretrained textual space: in semantically compressed domains, the encoder overfits local patterns and degrades the generalization capability from large-scale pretraining. In contrast, exclusively optimizing the visual branch preserves pretrained capability while adapting to domain-specific visual details.

\paragraph{Performance Degradation of Standard ISS.} As discussed in Section~\ref{sec:rethinking_methods}, conventional ISS imposes an ill-posed constraint by separating semantically near-identical images. Table~\ref{tab:issablation} confirms it causes consistent performance drops across nearly all models in vehicle and pedestrian domains.

\paragraph{Ablation of Hard Negative Mining.} We evaluate the HNM strategy by integrating it into our standard ITC baseline. Table~\ref{tab:ablation_hnm} shows that HNM yields only marginal gains and can even degrade retrieval performance, as its rigid decision boundaries cannot reliably distinguish genuine hard negatives from false negatives under such dense semantic overlap, potentially leading to dataset-specific overfitting.

\paragraph{Ablations of Loss Functions.}
As detailed in Table~\ref{tab:ablation_reordered}, we evaluate the contributions of the SASS and ISD components. SASS consistently improves retrieval performance over the baseline across all architectures, demonstrating the effectiveness of cross-modal soft-label supervision in alleviating false negatives. ISD further delivers additional gains, validating the complementary role of structural distillation in boosting domain-specific retrieval.

\section{Conclusion}

In this paper, we propose a systematic approach for domain-specific TBIR that encompasses benchmarking model performance in specific scenarios and improving cross-modal alignment under severe semantic compression. Specifically, we design the DSMM-TBIR data engine to construct the SecMM-TBIR benchmark. Integrated with our SAFT framework, it establishes a robust training and evaluation pipeline. Extensive results demonstrate that our approach substantially improves domain-specific TBIR performance and further enhances general-domain representation capabilities.

Future works may include extending this pipeline to more specialized domains, while exploring lightweight deployment on edge devices also remains an open research topic.

\clearpage
\bibliography{aaai2027}


\clearpage
\appendix
\section{Benchmark Details}

We construct SecMM-TBIR with our proposed DSMM-TBIR pipeline, with the Adaptive Data Pool Curation stage serving as the core component of our data preparation workflow. In this stage, we build the image and caption pools by aggregating multi-source data drawn from three complementary channels: self-collected domain-specific datasets, targeted web crawling of in-domain visual-textual pairs, and a curated collection of publicly available benchmarks, notably encompassing CUHK-PEDES, RSTPReid, ICFG-PEDES, SYNTH-PEDES, and CUHK-CompCars.

\textbf{Textual Branch.} We first analyze the text distribution of the collected data to derive corresponding statistical information, which are incorporated into prompts as detailed in Table~\ref{tab:Prompt_Pedestrian}. The resulting prompts are then fed into Qwen3 to generate comprehensive textual queries.

\textbf{Visual Branch.} Images are first encoded by DINOv3 to obtain visual embeddings. CGDS then applies K-means clustering to the pedestrian and vehicle image pools separately in the embedding space, with $\tilde{K}=100$ in each domain, followed by randomly sampling 100–1000 images from each cluster. Qwen3-VL subsequently filters the sampled images to retain only those relevant to the target object category.

\begin{figure}[!b]
    \centering
    \includegraphics[width=1.0\linewidth]{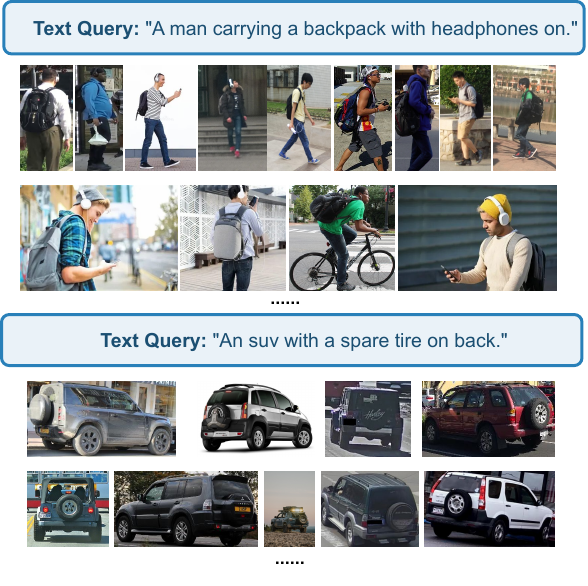}
    \caption{\textbf{More SecMM-TBIR Samples.}}
    \label{fig:more_viz}
\end{figure}

\textbf{Multi-Expert Collaborative Filtering.} A suite of universal multi-modal embedding models serve as expert retrievers, including Qwen3-VL-Embedding, Jina-v4, RZen-v2, and SigLIP. Each expert retrieves its top-$\hat{K}=1000$ candidates for a given query. We then construct the final set of potential matching pairs by taking the union of all retrieved results, which proceeds to annotation.

As shown in Figure \ref{fig:more_viz}, additional visualization examples of our SecMM-TBIR are provided for easier intuitive understanding, where each query is associated with multiple relevant visual candidates. 

\begin{figure}[!t]
    \centering
    \includegraphics[width=1.0\linewidth]{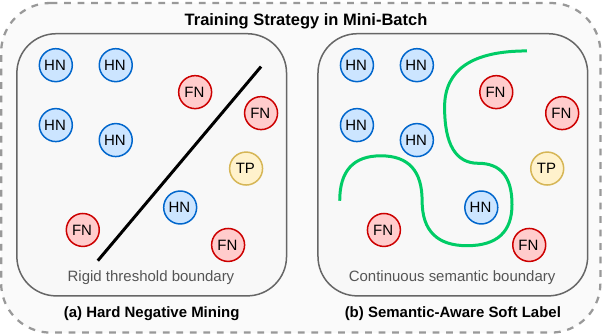}
    \caption{\textbf{Mini-batch Training Strategies.} Hard Negative Mining (HNM) enforces a rigid decision boundary, whereas SASS utilizes soft cross-modal distributions to preserve continuous semantic similarities.}
    \label{fig:classification}
\end{figure}

\begin{table}[!b]
    \centering
    \vskip 0.15in
    \begin{small}
    \begin{tabular}{lc}
    \toprule
    \multirow{2}{*}{\textbf{Hyperparameter}} & \textbf{Value} \\
                    & TinyCLIP, MobileCLIP, OpenCLIP \\
    \midrule
    % \pagebreak
    Input resolution  & $224^2$, $256^2$, $224^2$ \\
    Resize Crop Scale & [0.8, 1.0] \\
    Resize Crop Ratio & [0.2, 2.0] \\
    \midrule
    Training Iterations & 25k \\
    Batch Size & 128 \\
    \midrule
    Optimizer & AdamW \\
    Min LR & 5e-7 \\
    Max LR & 5e-6 \\
    Warmup Ratio & 0.1 \\
    LR Decay Schedule & Cosine Annealing \\
    Weight Decay Rate & 0.05 \\
    \midrule
    SASS Loss Weight $\alpha$ & 1 \\
    ISD Loss Weight $\beta$ & 0.75 \\
    \midrule
    SASS Teacher & UniME-V2-7B \\
    ISD Teacher  & UniME-V2-7B \\
    \bottomrule
    \end{tabular}
    \caption{\textbf{Hyperparameter Configurations for the Fine-Tuning Framework.}}
    \label{tab:hyperparameters}
    \end{small}
\end{table}

\section{Training Hyperparameters}
For completeness and reproducibility, we explicitly provide full supplementary implementation details on the hyperparameter configurations used in our main experiments. The corresponding hyperparameter settings for different experiments are fully listed in Table~\ref{tab:hyperparameters}.

We adopt a relatively small batch size and a conservative learning rate schedule to ensure stable optimization under severe semantic compression in domain-specific TBIR tasks. The scale and aspect ratio ranges for random resized cropping are adjusted to fit the typical visual profiles of pedestrian and vehicle targets. Concretely, the scale range is empirically fixed at $[0.8, 1.0]$ to effectively retain key foreground information, and the aspect ratio range is set to $[0.2, 2.0]$ to handle the varied shapes and orientations of pedestrians and vehicles in the real-world.

\begin{table}[!t]
\centering
\footnotesize
\renewcommand{\arraystretch}{1}
\setlength{\tabcolsep}{3pt}

\begin{tabular*}{\columnwidth}{@{\extracolsep{\fill}} l *{3}{c} }
    \toprule
    \multirow{3}{*}{\textbf{Model}} & \multicolumn{3}{c}{\textbf{Fashion200k}} \\
    \cmidrule(lr){2-4}
    & \multicolumn{1}{c}{R$@$1} & \multicolumn{1}{c}{R$@$5} & \multicolumn{1}{c}{R$@$10}\\
    \midrule
    
    % --- TinyCLIP-32M Group ---
    TinyCLIP-22M/32                 & 35.7 & 65.0 & 76.3 \\
    \quad + ITC                   & 39.4 & 66.8  &  77.0 \\
    \quad + CUSA                  & 39.2 & 68.4  & 77.2 \\
    \quad \textbf{+ SAFT (ours)}  & \textbf{39.7} & \textbf{68.6}  & \textbf{78.0} \\
    \addlinespace[3pt]
    \hline
    \addlinespace[3pt]
    
    % --- TinyCLIP-63M Group ---
    TinyCLIP-45M/32               & 42.2 & 71.0 & 80.2\\
    \quad + ITC                   & 44.3 & 71.5 & 80.7  \\
    \quad + CUSA                  & 44.6 & 71.3 & 81.8 \\
    \quad \textbf{+ SAFT (ours)}  & \textbf{45.7} & \textbf{73.7} & \textbf{82.8} \\
    \addlinespace[3pt]
    \hline
    \addlinespace[3pt]
    
    % --- MobileCLIP-S0 Group ---
    MobileCLIP-S0                   & 45.1 & 74.8 & 84.6 \\
    \quad + ITC                   & 52.7 & 83.9 & 91.9  \\
    \quad + CUSA                  & 52.1 & 83.2 & 91.4  \\
    \quad \textbf{+ SAFT (ours)}  & \textbf{55.3} & \textbf{84.3} & \textbf{92.3}  \\
    \addlinespace[3pt]
    \hline
    \addlinespace[3pt]
    
    % --- MobileCLIP-S1 Group ---
    MobileCLIP-S1                 & 53.7 & 81.7 & 89.4\\
    \quad + ITC                   & 60.8 & 88.6 & 94.7  \\
    \quad + CUSA                  & 59.8 & 88.1 & 94.6  \\
    \quad \textbf{+ SAFT (ours)}  & \textbf{61.5} & \textbf{88.7} & \textbf{95.3}  \\
    \addlinespace[3pt]
    \hline
    \addlinespace[3pt]
    
    % --- OpenCLIP-B/32 Group ---
    OpenCLIP-B/32                   & 38.3 & 67.5 & 77.3 \\
    \quad + ITC                   & 42.4 & 69.1 & 79.4  \\
    \quad + CUSA                  & 42.7 & 71.1 & 80.9  \\
    \quad \textbf{+ SAFT (ours)}  & \textbf{42.8} & \textbf{71.2} & \textbf{81.5}  \\
    \addlinespace[3pt]
    \hline
    \addlinespace[3pt]
    
    % --- OpenCLIP-B/16 Group ---
    OpenCLIP-B/16                   & 40.2 & 69.1 & 78.2 \\
    \quad + ITC                   & 44.7 & 71.1 & 81.8  \\
    \quad + CUSA                  & 45.2 & 73.8 & 82.2  \\
    \quad \textbf{+ SAFT (ours)}  & \textbf{46.3} & \textbf{74.2} & \textbf{82.7}  \\
    \bottomrule
\end{tabular*}
\caption{\textbf{Results on Fashion200K.}}
\label{tab:fashion_200k}
\end{table}

\section{More Detailed Results}

\paragraph{Domain Generalization on Fashion200K.} To further evaluate the domain generalization capability of our framework, we conduct additional experiments on the Fashion200K dataset, extending the evaluation scope from surveillance domain to commercial fashion scenarios. To adapt the benchmark to practical text-based image retrieval settings, we augment the original dataset by generating item descriptions using the Qwen3-VL together with Qwen-3.5, from which 2,000 of the resulting samples are selected to form the test set, while the remaining samples are used to form the training set.

Quantitative results are summarized in Table~\ref{tab:fashion_200k}. Our method
consistently outperforms all baselines by a significant margin
in this completely different domain. This validation confirms the broad applicability of our method beyond security-focused surveillance tasks and demonstrates its effectiveness and generalizability for domain-specific TBIR.

\paragraph{Compositional Reasoning Capability.} We further validate the compositional reasoning capability of our method on the ARO benchmark. The model is trained using the same training set described in Experiment Section. As reported in Table~\ref{tab:aro}, SAFT improves compositional reasoning performance across most evaluated models on both the Relation and Attribution subsets. While the gains vary across different models, the overall trend suggests that SAFT enhances the model's ability to capture fine-grained semantic structures and complex multi-modal compositional relations.

\begin{table}[!t]
\centering
\footnotesize
\renewcommand{\arraystretch}{1}
\setlength{\tabcolsep}{3pt}

\begin{tabular*}{\columnwidth}{@{\extracolsep{\fill}} l *{2}{c} }
    \toprule
    \multirow{3}{*}{\textbf{Model}} & \multicolumn{2}{c}{\textbf{ARO (Acc. $\uparrow$)}} \\
    \cmidrule(lr){2-3}
    & \multicolumn{1}{c}{\textbf{Relation}} & \multicolumn{1}{c}{\textbf{Attribution}} \\
    \midrule
    
    % --- TinyCLIP-32M Group ---
    TinyCLIP-22M/32                 & 48.2 & 56.0  \\
    \quad + ITC                   & \textbf{51.7} & \textbf{60.6} \\
    \quad + CUSA                  & 50.8 & 57.2 \\
    \quad \textbf{+ SAFT (ours)}  & 51.3 & 60.0 \\
    % \cmidrule(lr){1-3}
    \addlinespace[3pt]
    \hline
    \addlinespace[3pt]
    
    % --- TinyCLIP-63M Group ---
    TinyCLIP-45M/32               & 49.8 & 59.1 \\
    \quad + ITC                   & 51.2 & 61.2  \\
    \quad + CUSA                  & 51.8 & 61.6 \\
    \quad \textbf{+ SAFT (ours)}  & \textbf{51.9} & \textbf{61.9} \\
    \addlinespace[3pt]
    \hline
    \addlinespace[3pt]
    
    % --- MobileCLIP-S0 Group ---
    MobileCLIP-S0                   & 49.5 & 62.6 \\
    \quad + ITC                   & 51.1 & 63.8  \\
    \quad + CUSA                  & 50.9 & 63.7  \\
    \quad \textbf{+ SAFT (ours)}  & \textbf{53.5} & \textbf{65.7}  \\
    \addlinespace[3pt]
    \hline
    \addlinespace[3pt]
    
    % --- MobileCLIP-S1 Group ---
    MobileCLIP-S1                   & 50.2 & 65.7\\
    \quad + ITC                   & 50.6 & 65.0  \\
    \quad + CUSA                  & 50.6 & 65.4  \\
    \quad \textbf{+ SAFT (ours)}  & \textbf{53.0} & \textbf{67.2}  \\
    \addlinespace[3pt]
    \hline
    \addlinespace[3pt]
    
    % --- OpenCLIP-B/32 Group ---
    OpenCLIP-B/32                 & 49.8 & 58.5 \\
    \quad + ITC                   & 50.9 & \textbf{60.8}  \\
    \quad + CUSA                  & 50.6 & 60.0  \\
    \quad \textbf{+ SAFT (ours)}  & \textbf{51.0} & \textbf{60.8}  \\
    \addlinespace[3pt]
    \hline
    \addlinespace[3pt]
    
    % --- OpenCLIP-B/16 Group ---
    OpenCLIP-B/16                 & 46.6 & 56.8\\
    \quad + ITC                   & 49.9 & 59.8  \\
    \quad + CUSA                  & 50.6 & 60.2  \\
    \quad \textbf{+ SAFT (ours)}  & \textbf{51.2} & \textbf{61.0}  \\
    \bottomrule
\end{tabular*}
\caption{\textbf{Evaluation on ARO Benchmark.}}
\label{tab:aro}
\end{table}

\begin{table*}[t]
\centering
\small
\renewcommand{\arraystretch}{1.25}

\begin{tabular}{p{1.5cm}p{2.5cm}p{12.5cm}}
\toprule
\textbf{Task} & \textbf{Prompt Category} & \textbf{Prompt} \\
\midrule

\multirow{2}{*}[0.6em]{Pedestrian}
&
System
&
You are a data engineer responsible for constructing a text-image retrieval dataset. 
Based on the following requirements, please generate a batch of English text retrieval queries centered around "pedestrians/people" for security camera scenarios.

\vspace{0.5em}

\textbf{[Task Scenario]}
\begin{itemize}
    \item All outputs must be English query sentences.
    \item The content of the queries must center around "people/pedestrians."
    \item The tone of the queries should match the realistic search behavior of security system users.
\end{itemize}

\vspace{0.5em}

\textbf{[Template Structure]} \newline
All query sentences must adhere to the following structure: \newline
\texttt{[person] + [specific action and/or descriptor] + (optional) second or third complementary description}

\vspace{0.5em}

\textbf{[Stylistic Requirements]}
\begin{itemize} %[nosep, leftmargin=*]
    \item \textbf{Sentence Length:} Approximately 5-15 English words (MS-COCO caption style, with natural variations in query length).
    \item \textbf{Tone \& Style:} Natural, concise, and aligned with real-world surveillance search queries, avoiding overly literary or descriptive language.
    \item \textbf{Diversity:} Cover a wide range of human attributes, including actions, appearance, clothing, carried objects, and characteristics.
    \item \textbf{Safety Guardrails:} Generation of any content related to danger, violence, pornography, weapons, or illegal activities is strictly prohibited.
\end{itemize}

\vspace{0.5em}

\textbf{[Semantic Background]} \newline
Common nouns, verbs, adjectives, and adverbs related to people have been statistically extracted from relevant datasets. Please randomly select and combine elements from these semantic scopes to ensure high diversity in the generated queries.

\vspace{0.5em}

\textbf{[Output Format]}
\begin{itemize} %[nosep, leftmargin=*]
    \item Output multiple English queries.
    \item Each query should be on a new line, without any numbering or bullet points.
    \item Do not output any explanations or extra commentary; output the queries only.
\end{itemize}

\vspace{0.5em}

\textbf{[Output Examples]} \newline
A little boy sitting alone on a motorcycle. \newline
A man rides a motorcycle down an empty street next to houses. \newline
The woman in the kitchen is holding a huge pan. \newline
A man standing in a kitchen while closing a cupboard door. \newline
Man in motorcycle leathers standing in front of a group of bikes. \newline
Two people riding on a moped with a bus in the next lane. \newline
A girl in bikini standing with surfboard on the beach. \newline
A man holds a hot dog while several people walk in the background. \newline
A man with a racket on a tennis court. \newline
A woman in a pink jacket and printed pants. \newline
A man with a red shirt is holding a cell phone up to his ear.
\\
\cmidrule(l){2-3}

&
User
&
Please randomly select from the following keywords and generate \texttt{[part\_num]} pedestrian-related descriptions (try to use most of the words):\newline
Nouns: [keywords\_noun\_part\_str]\newline
Verbs: [keywords\_verb\_part\_str]\newline
Adjectives: [keywords\_adj\_part\_str]\newline
Please start generating queries:\newline
\\
\bottomrule
\end{tabular}
\caption{\textbf{Prompt Engineering in Distribution-Aware Prompting (DAP).}}
\label{tab:Prompt_Pedestrian}
\end{table*}

\end{document}